\documentclass{article}

\usepackage[numbers]{natbib}
\usepackage[preprint]{neurips_2022}

\usepackage[dvipsnames]{xcolor}         
\definecolor{linkColor}{rgb}{0.18,0.39,0.62}
\usepackage[utf8]{inputenc} 
\usepackage[T1]{fontenc}    
\usepackage[colorlinks=true,linkcolor=linkColor,citecolor=linkColor,filecolor=linkColor,urlcolor=linkColor]{hyperref}       
\usepackage{url}            
\usepackage{booktabs}       
\usepackage{amsfonts}       
\usepackage{nicefrac}       
\usepackage{microtype}      

\RequirePackage{algorithm}
\RequirePackage{algorithmic}

\usepackage{multirow}
\usepackage{amsmath}
\usepackage{capt-of}
\usepackage{tabularx}
\usepackage{epsfig}
\usepackage{amssymb}
\usepackage{amsfonts}
\usepackage{booktabs}
\usepackage{scalerel}
\usepackage[inline]{enumitem}
\usepackage{listings}
\usepackage{varwidth}
\usepackage[export]{adjustbox}
\usepackage{tikz}
\usetikzlibrary{tikzmark}

\usepackage{stmaryrd}
\usepackage{bbm}
\usepackage{wrapfig}
\usepackage{pifont}

\newcommand{\tabincell}[2]{\begin{tabular}{@{}#1@{}}#2\end{tabular}}

\newcommand{\sptk}[1]{\texttt{[#1]}}

\definecolor{deepblue}{rgb}{0,0,0.5}
\definecolor{officeblue}{RGB}{0,102,204}
\definecolor{deepred}{rgb}{0.6,0,0}
\definecolor{deepgreen}{rgb}{0,0.5,0}
\definecolor{mybrickred}{RGB}{182,50,28}

\definecolor{fillcolor}{RGB}{216,217,252}

\usepackage{amsmath,amsfonts,bm}

\def\eqref#1{equation~\ref{#1}}

\def\1{\bm{1}}

\DeclareMathAlphabet{\mathsfit}{\encodingdefault}{\sfdefault}{m}{sl}
\SetMathAlphabet{\mathsfit}{bold}{\encodingdefault}{\sfdefault}{bx}{n}

\usepackage{pifont}
\newcommand{\xmark}{{\color{black}\ding{55}}}%

\newcommand\our{\textsc{BEiT-3}}

\newcommand\multiway{Multiway Transformers}
\newcommand\beit{\textsc{BEiT}}

\newcommand{\boxAP}{{AP$^\text{box}$}}
\newcommand{\maskAP}{{AP$^\text{mask}$}}

\title{Image as a Foreign Language: \beit{} Pretraining for All Vision and Vision-Language Tasks}

\author{%
Wenhui Wang\thanks{~Equal contribution. $\dagger$ Corresponding author.}, ~~Hangbo Bao\footnotemark[1], ~~Li Dong\footnotemark[1], ~~Johan Bjorck,~~Zhiliang Peng,~~Qiang Liu \\\textbf{Kriti Aggarwal,}~~\textbf{Owais Khan Mohammed,}~~\textbf{Saksham Singhal,}~~\textbf{Subhojit Som,}~~\textbf{Furu Wei}$^\dagger$ \\
Microsoft Corporation \\
\url{https://aka.ms/beit-3}
}

\begin{document}

\maketitle

\vspace{-6mm}
\begin{abstract}
A big convergence of language, vision, and multimodal pretraining is emerging. In this work, we introduce a general-purpose multimodal foundation model \textbf{\our{}}, which achieves state-of-the-art transfer performance on both vision and vision-language tasks. Specifically, we advance the big convergence from three aspects: backbone architecture, pretraining task, and model scaling up. We introduce \multiway{} for general-purpose modeling, where the modular architecture enables both deep fusion and modality-specific encoding. Based on the shared backbone, we perform masked ``language'' modeling on images ({\textbf{Imglish}}), texts (English), and image-text pairs (``parallel sentences'') in a unified manner. Experimental results show that \our{} obtains state-of-the-art performance on object detection (COCO), semantic segmentation (ADE20K), image classification (ImageNet), visual reasoning (NLVR2), visual question answering (VQAv2), image captioning (COCO), and cross-modal retrieval (Flickr30K, COCO).
\end{abstract}

\vspace{-5mm}
\begin{figure}[h]
\begin{center}
\begin{tabular}{c}
\includegraphics[width=0.9\textwidth]{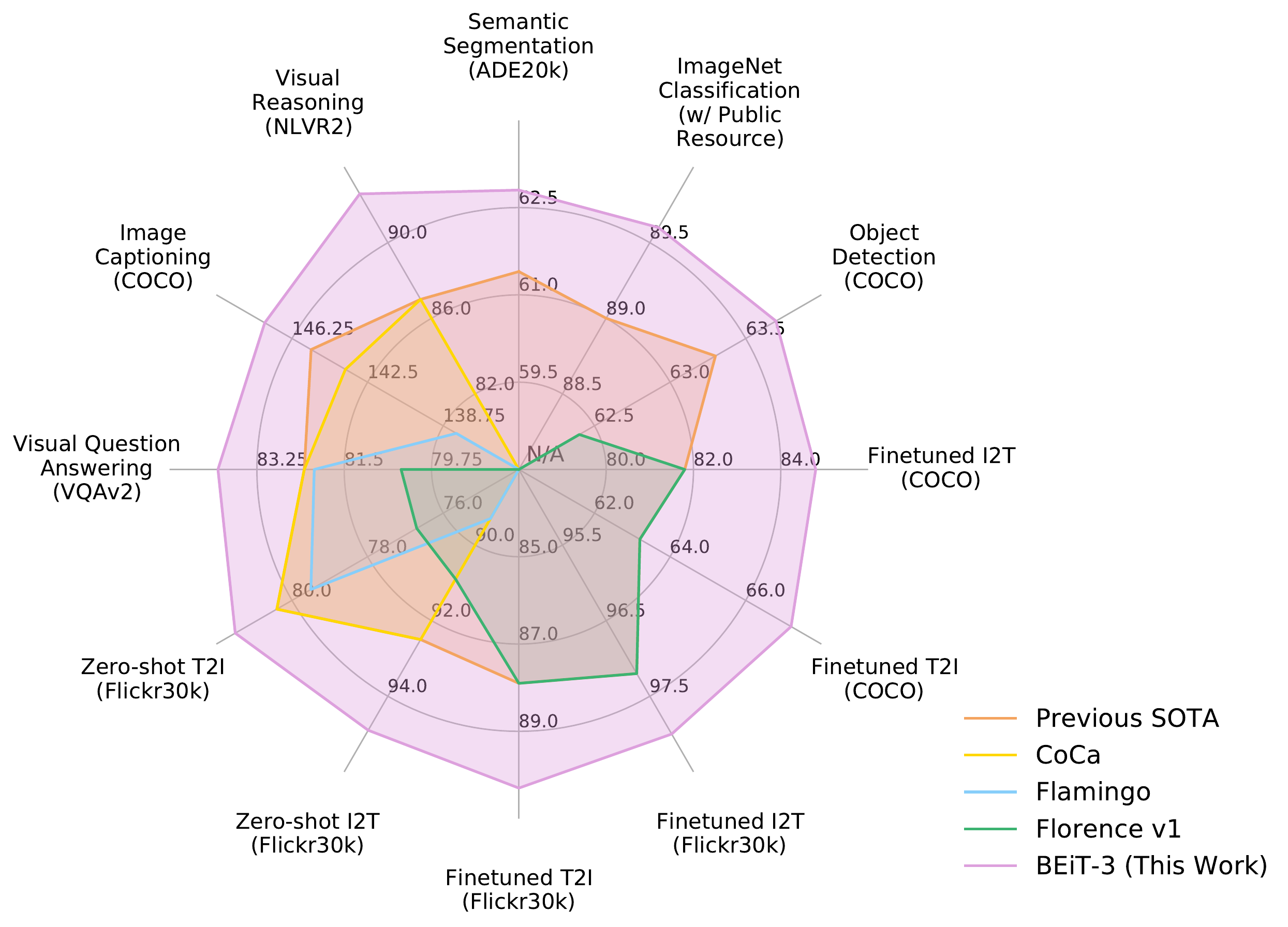}
\end{tabular}
\end{center}
\vspace{-2mm}
\caption{\our{} achieves state-of-the-art performance on a broad range of tasks compared with other customized or foundation models. I2T/T2I is short for image-to-text/text-to-image retrieval. 
}
\label{fig:radar_compare}
\end{figure}

\begin{table*}
\centering
\small
\begin{tabular}{@{}llllll@{}}
\toprule
\bf Category & \bf Task & \bf Dataset & \bf Metric & \bf Previous SOTA & \bf \our{} \\
\midrule
\multirow{3}{*}{Vision} & Semantic Segmentation & ADE20K & mIoU & 61.4 (FD-SwinV2) & \bf 62.8 (\textcolor{Green}{+1.4}) \\
\cmidrule{2-6}
& Object Detection & COCO & AP & 63.3 (DINO) & \bf 63.7
(\textcolor{Green}{+0.4}) \\
& Instance Segmentation & COCO & AP & 54.7 (Mask DINO) & \bf 54.8
(\textcolor{Green}{+0.1}) \\
\cmidrule{2-6}
& Image Classification & ImageNet$\dag$ & Top-1 acc. & 89.0 (FD-CLIP) & \bf 89.6 (\textcolor{Green}{+0.6}) \\
\midrule
\multirow{6}{*}{Vision-Language} & Visual Reasoning & NLVR2 & Acc. & 87.0 (CoCa) & \bf 92.6 (\textcolor{Green}{+5.6}) \\
\cmidrule{2-6}
& Visual QA & VQAv2 & VQA acc. & 82.3 (CoCa) & \bf 84.0 (\textcolor{Green}{+1.7}) \\
\cmidrule{2-6}
& Image Captioning & COCO$\ddag$ & CIDEr & 145.3 (OFA) & \bf 147.6 (\textcolor{Green}{+2.3}) \\
\cmidrule{2-6}
& \multirow{2}{*}{Finetuned Retrieval} & COCO & \multirow{2}{*}{R@1} & 72.5 (Florence) & \bf 76.0 (\textcolor{Green}{+3.5}) \\
& & Flickr30K & & 92.6 (Florence) & \bf 94.2 (\textcolor{Green}{+1.6}) \\
\cmidrule{2-6}
& Zero-shot Retrieval & Flickr30K & R@1 & 86.5 (CoCa) & \bf 88.2 (\textcolor{Green}{+1.7}) \\
\bottomrule
\end{tabular}
\caption{Overview of \our{} results on various vision and vision-language benchmarks. 
We compare with previous state-of-the-art models, including FD-SwinV2~\citep{fd-swin}, DINO~\citep{dino-od}, Mask DINO~\citep{dino-od}, FD-CLIP~\citep{fd-swin}, CoCa~\citep{coca}, OFA~\citep{ofa}, Florence~\citep{florence}.
We report the average of top-$1$ image-to-text and text-to-image results for retrieval tasks.
``$\dag$'' indicates ImageNet results only using publicly accessible resources.
``$\ddag$'' indicates image captioning results without CIDEr optimization.
}
\label{tab:presota_comparision}
\end{table*}


\section{Introduction: The Big Convergence}
\label{sec:intro}

Recent years have featured a trend toward the big convergence of language~\citep{gpt,bert,unilm}, vision~\citep{beit,beitv2}, and multimodal~\citep{vlmo,clip,coca} pretraining.
By performing large-scale pretraining on massive data, we can easily transfer the models to various downstream tasks.
It is appealing that we can pretrain a general-purpose foundation model that handles multiple modalities.
In this work, we advance the convergence trend for vision-language pretraining from the following three aspects.

First, the success of Transformers~\citep{transformer} is translated from language to vision~\citep{vit} and multimodal~\citep{vilt,vlmo} problems.
The unification of network architectures enables us to seamlessly handle multiple modalities.
For vision-language modeling, there are various ways to apply Transformers due to the different natures of downstream tasks.
For example, the dual-encoder architecture is used for efficient retrieval~\citep{clip}, encoder-decoder networks for generation tasks~\citep{simvlm}, and the fusion-encoder architecture for image-text encoding~\citep{vilt}.
However, most foundation models have to manually convert the end-task formats according to the specific architectures. Moreover, the parameters are usually not effectively shared across modalities.
In this work, we adopt \multiway{}~\citep{vlmo} for general-purpose modeling, i.e., one unified architecture shared for various downstream tasks. The modular network also comprehensively considers modality-specific encoding and cross-modality fusion.

Second, the pretraining task based on masked data modeling has been successfully applied to various modalities, such as texts~\citep{bert}, images~\citep{beit,beitv2}, and image-text pairs~\citep{vlbeit}.
Current vision-language foundation models usually multitask other pretraining objectives (such as image-text matching), rendering scaling-up unfriendly and inefficient.
In contrast, we only use one pretraining task, i.e., mask-then-predict, to train a general-purpose multimodal foundation model.
By regarding the image as a foreign language (i.e., \textit{Imglish}), we handle texts and images in the same manner without fundamental modeling differences.
Consequentially, image-text pairs are utilized as ``parallel sentences'' in order to learn the alignments between modalities.
We also show that the simple yet effective method learns strong transferable representations, achieving state-of-the-art performance on both vision and vision-language tasks.
The prominent success demonstrates the superiority of generative pretraining~\citep{bert,beit}.

Third, scaling up the model size and data size universally improves the generalization quality of foundation models, so that we can transfer them to various downstream tasks.
We follow the philosophy and scale up the model size to billions of parameters.
Moreover, we scale up the pretraining data size in our experiments while only using publicly accessible resources for academic reproducibility.
Although without using any private data, our method outperforms state-of-the-art foundation models that rely on in-house data by a decent margin.
In addition, the scaling up benefits from treating images as a foreign language, as we can directly reuse the pipeline developed for large-scale language model pretraining.

In this work, we take advantage of the above ideas to pretrain a general-purpose multimodal foundation model \our{}.
We pretrain a Multiway Transformer by performing masked data modeling on images, texts, and image-text pairs.
During pretraining, we randomly mask some proportion of text tokens or image patches. The self-supervised learning objective is to recover the original tokens (i.e., text tokens, or visual tokens) given corrupted inputs.
The model is general-purpose in the sense that it can be repurposed for various tasks regardless of input modalities, or output formats.

As shown in Figure~\ref{fig:radar_compare} and Table~\ref{tab:presota_comparision}, \our{} achieves state-of-the-art transfer performance across a broad range of vision and vision-language tasks.
We evaluate \our{} on extensive downstream tasks and datasets, i.e., object detection (COCO), instance segmentation (COCO), semantic segmentation (ADE20K), image classification (ImageNet), visual reasoning (NLVR2), visual question answering (VQAv2), image captioning (COCO), and cross-modal retrieval (Flickr30K, COCO).
Specifically, our model outperforms previous strong foundation models~\citep{coca,flamingo,florence} despite that we only use public resources for pretraining and finetuning.
The model also obtains better results than specialized models.
Moreover, \our{} not only performs well on vision-language tasks but also on vision tasks (such as object detection, and semantic segmentation).

\section{\our{}: A General-Purpose Multimodal Foundation Model}
\label{sec:methods}

\begin{figure}[t]
\begin{center}
\begin{tabular}{c}
\includegraphics[width=0.86\textwidth]{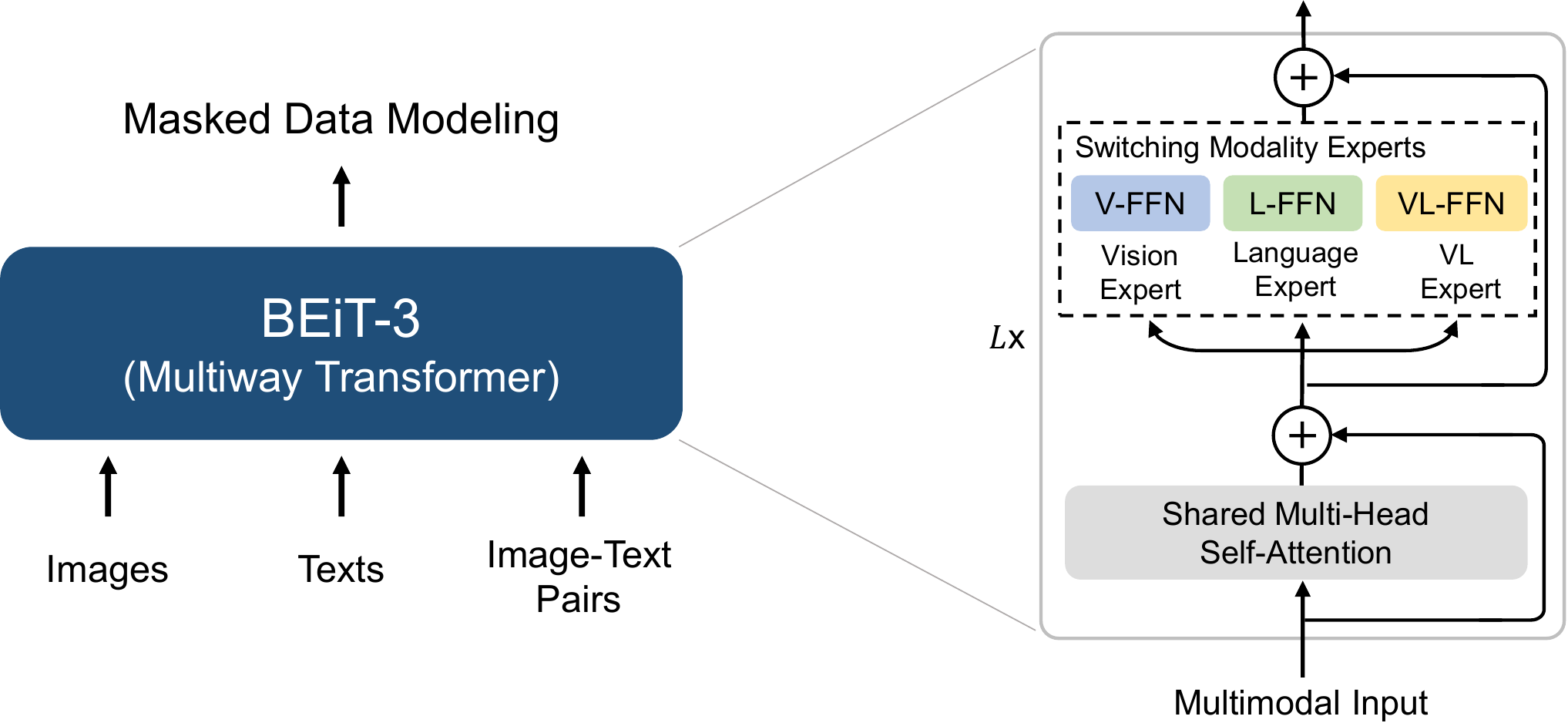}
\end{tabular}
\end{center}
\caption{
Overview of \our{} pretraining.
We perform masked data modeling on monomodal (i.e., images, and texts) and multimodal (i.e., image-text pairs) data with a shared Multiway Transformer as the backbone network.
}
\label{fig:overview}
\end{figure}

\begin{figure}[t]
\begin{center}
\begin{tabular}{c}
\includegraphics[width=\textwidth]{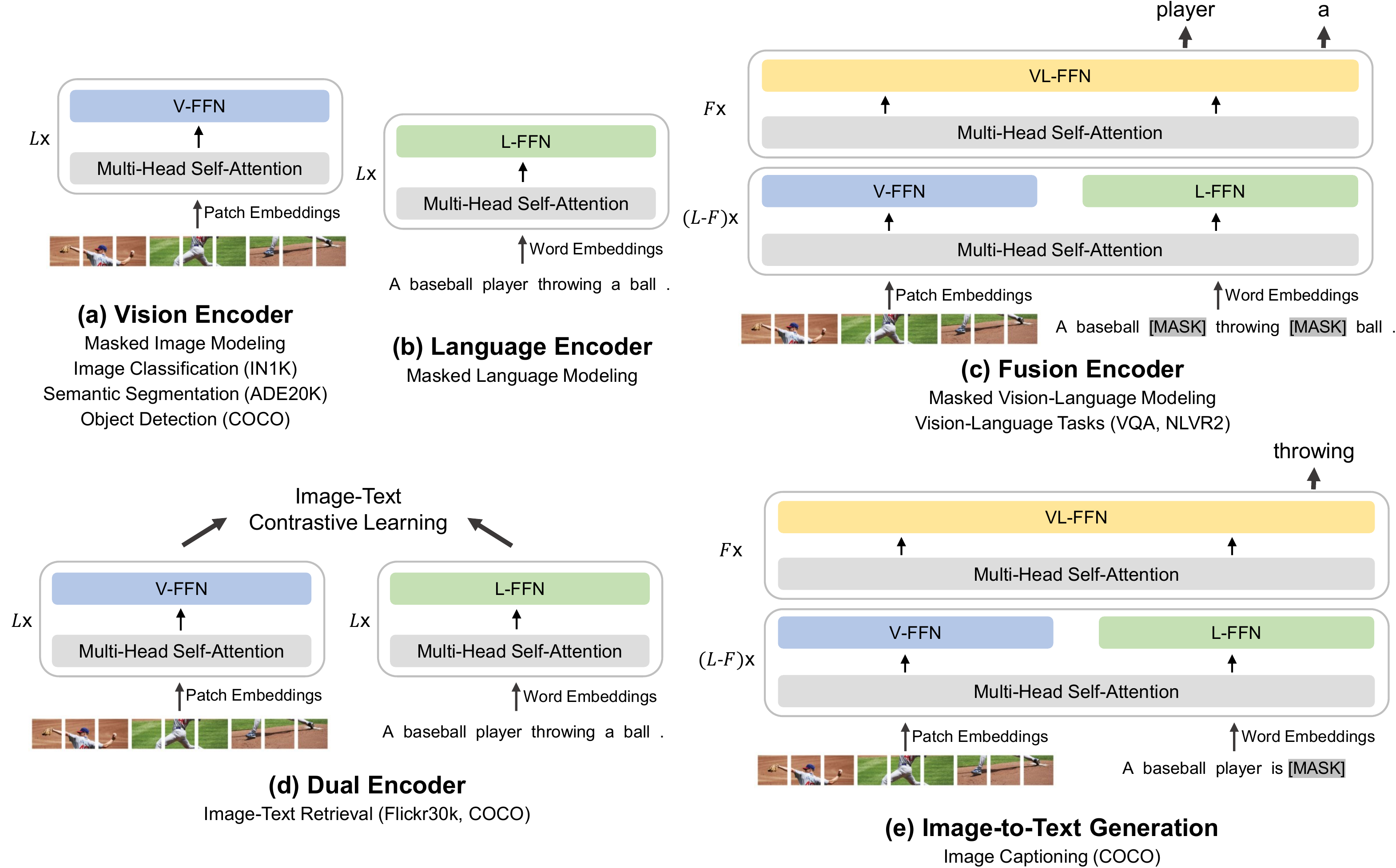}
\end{tabular}
\end{center}
\caption{
\our{} can be transferred to various vision and vision-language downstream tasks.
With a shared Multiway Transformer, we can reuse the model as (a)(b) vision or language encoders; (c) fusion encoders that jointly encode image-text pairs for deep interaction; (d) dual encoders that separately encode modalities for efficient retrieval; (e) sequence-to-sequence learning for image-to-text generation.
}
\label{fig:multiway}
\end{figure}

As shown in Figure~\ref{fig:overview}, \our{} is pretrained by masked data modeling on monomodal and multimodal data, using a shared Multiway Transformer network.
The model can be transferred to various vision and vision-language downstream tasks.

\subsection{Backbone Network: \multiway{}}

We use \multiway{}~\citep{vlmo} as the backbone model to encode different modalities.
As shown in Figure~\ref{fig:overview}, each Multiway Transformer block consists of a shared self-attention module, and a pool of feed-forward networks (i.e., modality experts) used for different modalities.
We route each input token to the experts depending on its modality.
In our implementation, each layer contains a vision expert and a language expert. Moreover, the top three layers have vision-language experts designed for fusion encoders.
Refer to Figure~\ref{fig:multiway} (a)(b)(c) for more detailed modeling layouts.
Using a pool of modality experts encourages the model to capture more modality-specific information.
The shared self-attention module learns the alignment between different modalities and enables deep fusion for multimodal (such as vision-language) tasks.

As shown in Figure~\ref{fig:multiway}, the unified architecture enables \our{} to support a wide range of downstream tasks.
For example, \our{} can be used as an image backbone for various vision tasks, including image classification, object detection, instance segmentation, and semantic segmentation.
It can also be finetuned as a dual encoder for efficient image-text retrieval, and a fusion model for multimodal understanding and generation tasks.

\subsection{Pretraining Task: Masked Data Modeling}

We pretrain \our{} via a unified masked data modeling~\citep{vlbeit} objective on monomodal (i.e., images, and texts) and multimodal data (i.e., image-text pairs).
During pretraining, we randomly mask some percentage of text tokens or image patches and train the model to recover the masked tokens.
The unified mask-then-predict task not only learns representations but also learns the alignment of different modalities.
Specifically, text data is tokenized by a SentencePiece tokenizer~\citep{sentencepiece}.
Image data is tokenized by the tokenizer of \beit{} v2~\citep{beitv2} to obtain the discrete visual tokens as the reconstructed targets.
We randomly mask $15$\% tokens of monomodal texts and $50$\% tokens of texts from image-text pairs.
For images, we mask $40$\% of image patches using a block-wise masking strategy as in \beit{}~\citep{beit,beitv2}.

We only use one pretraining task, which makes the training process scaling-up friendly. In contrast, previous vision-language models~\citep{oscar,vinvl,vilt,albef,vlmo,blip,coca} usually employ multiple pretraining tasks, such as image-text contrast, image-text matching, and word-patch/region alignment.
We show that a much smaller pretraining batch size can be used with the mask-then-predict task.
In comparison, contrastive-based models~\cite{clip,align,florence,coca} usually need a very large batch size\footnote{For example, CoCa~\citep{coca} uses $65$k batch size, CLIP~\citep{clip} uses $32$k batch size, and Florence~\citep{florence} uses $24$k batch size. \our{} uses a much smaller $6$k batch size for pretraining.} for pretraining, which brings more engineering challenges, such as GPU memory cost.

\subsection{Scaling Up: \our{} Pretraining}

\paragraph{Backbone Network}

\our{} is a giant-size foundation model following the setup of ViT-giant~\citep{scaling:vit}.
As shown in Table~\ref{tab:model_config}, the model consists of a $40$-layer Multiway Transformer with $1408$ hidden size, $6144$ intermediate size, and $16$ attention heads.
All layers contain both vision experts and language experts.
Vision-language experts are also employed in the top three Multiway Transformer layers.
The self-attention module is shared across different modalities.
\our{} consists of $1.9$B parameters in total, including $692$M parameters for vision experts, $692$M parameters for language experts, $52$M parameters for vision-language experts, and $317$M parameters for the shared self-attention module.
Notice that only vision-related parameters (i.e., comparable size as ViT-giant; about 1B) are activated when the model is used as a vision encoder.

\begin{table*}
\centering
\small
\begin{tabular}{@{}lcccccccc@{}}
\toprule
\multirow{2}{*}{\bf Model} & \multirow{2}{*}{\bf \#Layers} & \multirow{2}{*}{\bf \tabincell{c}{Hidden \\ Size}} & \multirow{2}{*}{\bf \tabincell{c}{MLP \\ Size}} & \multicolumn{5}{c}{\bf \#Parameters} \\
\cmidrule(lr){5-9}
 & & & & \bf V-FFN & \bf L-FFN & \bf VL-FFN & \bf Shared Attention & \bf Total \\
\midrule
\our{} & 40 & 1408 & 6144 & 692M & 692M & 52M & 317M & 1.9B \\
\bottomrule
\end{tabular}
\caption{Model configuration of \our{}. The architecture layout follows ViT-giant~\citep{scaling:vit}.}
\label{tab:model_config}
\end{table*}

\begin{table*}
\centering
\small
\begin{tabular}{@{}lll@{}}
\toprule
\bf Data & \bf Source & \bf Size \\
\midrule
Image-Text Pair & CC12M, CC3M, SBU, COCO, VG & 21M pairs \\
Image & ImageNet-21K & 14M images \\
Text & English Wikipedia, BookCorpus, OpenWebText, CC-News, Stories & 160GB documents \\
\bottomrule
\end{tabular}
\caption{Pretraining data of \our{}.
All the data are academically accessible.
}
\label{tab:pretraining_data}
\end{table*}

\paragraph{Pretraining Data}
\our{} is pretrained on both monomodal and multimodal data shown in Table~\ref{tab:pretraining_data}.
For multimodal data, there are about $15$M images and $21$M image-text pairs collected from five public datasets: Conceptual 12M (CC12M)~\citep{cc12m}, Conceptual Captions (CC3M)~\citep{gcc}, SBU Captions (SBU)~\citep{sbu}, COCO~\citep{coco} and Visual Genome (VG)~\citep{vg}.
For monomodal data, we use $14$M images from ImageNet-21K and $160$GB text corpora~\citep{unilm2} from English Wikipedia, BookCorpus~\citep{bookcorpus}, OpenWebText\footnote{\url{http://skylion007.github.io/OpenWebTextCorpus}}, CC-News~\citep{roberta}, and Stories~\citep{stories_data}.

\paragraph{Pretraining Settings}
We pretrain \our{} for $1$M steps.
Each batch contains $6144$ samples in total, including $2048$ images, $2048$ texts and $2048$ image-text pairs.
The batch size is much smaller than contrastive models~\citep{clip,align,coca}.
\our{} uses $14 \times 14$ patch size and is pretrained at resolution $224 \times 224$.
We use the same image augmentation as in \beit{}~\citep{beit}, including random resized cropping, horizontal flipping, and color jittering~\citep{coloraug}.
A SentencePiece tokenizer~\citep{sentencepiece} with $64$k vocab size is employed to tokenize the text data.
We use the AdamW~\citep{adamw} optimizer with $\beta_1=0.9$, $\beta_2=0.98$ and $\epsilon=$1e-6 for optimization.
We use a cosine learning rate decay scheduler with a peak learning rate of 1e-3 and a linear warmup of $10$k steps.
The weight decay is $0.05$.
Stochastic depth~\citep{drop_path} with a rate of $0.1$ is used.
The BEiT initialization algorithm\footnote{We first randomly initialize the parameters within a small range, e.g., $[-0.02, 0.02]$. Next, we rescale the $l$-th Transformer layer's output matrices (i.e., the last linear projection within each sublayer) of self-attention and FFN by $\frac{1}{\sqrt{2l}}$.}~\citep{beit} is used to stabilize Transformer training.

\section{Experiments on Vision and Vision-Language Tasks}
\label{sec:exps}

We extensively evaluate \our{} on major public benchmarks for both vision-language and vision tasks.
Table~\ref{tab:presota_comparision} presents the overview of results. 
\our{} obtains state-of-the-art performance on a wide range of vision and vision-language tasks.

\subsection{Vision-Language Downstream Tasks}

We evaluate the capabilities of \our{} on the widely used vision-language understanding and generation benchmarks, including visual question answering~\citep{vqa}, visual reasoning~\citep{nlvr2}, image-text retrieval~\citep{flickr30k,coco}, and image captioning~\citep{coco}.

\begin{table*}[t]
\centering
\begin{tabular}{@{}lcccccccc@{}}
\toprule
\multirow{2}{*}{\bf Model} & \multicolumn{2}{c}{\bf VQAv2} & \multicolumn{2}{c}{\bf NLVR2} & \multicolumn{4}{c}{\bf COCO Captioning} \\
\cmidrule(lr){2-3} \cmidrule(lr){4-5} \cmidrule(lr){6-9}
 & test-dev & test-std & dev & test-P & B@4 & M & C & S \\
\midrule
Oscar~\citep{oscar} & 73.61 & 73.82 & 79.12 & 80.37 & 37.4 & 30.7 & 127.8 & 23.5 \\
VinVL~\citep{vinvl} & 76.52 & 76.60 & 82.67 & 83.98 & 38.5 & 30.4 & 130.8 & 23.4 \\
ALBEF~\citep{albef} & 75.84 & 76.04 & 82.55 & 83.14 & - & - & - & - \\
BLIP~\citep{blip} & 78.25 & 78.32 & 82.15 & 82.24 & 40.4 & - & 136.7 & - \\
SimVLM~\citep{simvlm} & 80.03 & 80.34 & 84.53 & 85.15 & 40.6 & 33.7 & 143.3 & \bf 25.4 \\
Florence~\citep{florence} & 80.16 & 80.36 & - & - & - & - & - & - \\
OFA~\citep{ofa} & 82.00 & 82.00 & - & - & 43.9 & 31.8 & 145.3 & 24.8 \\
Flamingo~\citep{flamingo} & 82.00 & 82.10 & - & - & - & - & 138.1 & - \\
CoCa~\citep{coca} & 82.30 & 82.30 & 86.10 & 87.00 & 40.9 & \bf 33.9 & 143.6 & 24.7 \\
\midrule
\bf \our{} & \bf 84.19 & \bf 84.03 & \bf 91.51 & \bf 92.58 & \bf 44.1 & 32.4 & \bf 147.6 & \bf 25.4 \\
\bottomrule
\end{tabular}
\caption{Results of visual question answering, visual reasoning, and image captioning tasks.
We report \textit{vqa-score} on VQAv2 test-dev and test-standard splits, accuracy for NLVR2 development set and public test set (test-P).
For COCO image captioning, we report BLEU@4 (B@4), METEOR (M), CIDEr (C), and SPICE (S) on the Karpathy test split.
For simplicity, we report captioning results without using CIDEr optimization.
}
\label{tbl:results:vqa_nlvr2_captioning}
\end{table*}

\paragraph{Visual Question Answering (VQA)} 

The task requires the model to answer natural language questions about input images.
Following previous work~\citep{bottom_up_attn,vinvl,vilt}, we conduct finetuning experiments on the VQA v2.0 dataset~\citep{vqa} and formulate the task as a classification problem.
The model is trained to predict answers from the $3129$ most frequent answer candidates in the training set.
\our{} is finetuned as a fusion encoder to model deep interactions of images and questions for the VQA task.
We concatenate the embeddings of a given question and an image, and then feed the input embeddings into \multiway{} to jointly encode the image-question pair.
The final pooled output is fed into a classifier layer to predict the answer.
The results are present in Table~\ref{tbl:results:vqa_nlvr2_captioning}, \our{} outperforms all previous models by a large margin (more than $1.7$ points), pushing the state of the art to $84.03$ with a single model.

\paragraph{Visual Reasoning}
The task needs models to perform joint reasoning about images and natural language descriptions.
We evaluate the model on the popular NLVR2~\citep{nlvr2} benchmark, which is to determine whether a textual description is true about a pair of images.
Following previous work~\citep{vinvl,vilt}, we construct two image-text pairs based on the triplet input.
We finetune \our{} as a fusion encoder to jointly encode the image-text pairs.
The final pooled outputs of the two pairs are concatenated and then fed into a classifier layer to predict the label. 
As shown in Table~\ref{tbl:results:vqa_nlvr2_captioning}, \our{} achieves a new state-of-the-art result for visual reasoning, outperforming CoCa by about $5.6$ points. 
The performance on NLVR2 reaches above $90$\% for the first time.

\paragraph{Image Captioning}

The task aims to generate a natural language caption for the given image.
We use the COCO~\citep{coco} benchmark, finetune and evaluate the model on Karpathy split~\citep{karpathysplit}. 
Following \textsc{UniLM}~\citep{unilm} and s2s-ft~\citep{s2s-ft}, \our{} is used as a conditional generation model via masked finetuning.
To be more specific, a special self-attention mask is employed for the image captioning task.
Image tokens (i.e., image patches) can only attend to each other bidirectionally within the image sequence.
Tokens of the caption can attention to image tokens, their leftward caption tokens, and themselves.
During finetuning, we randomly mask some percentage of caption tokens.
The model is trained to recover these tokens based on the clues of the image and its leftward caption context.
We also mask the special boundary token \sptk{SEP} to help the model learn to terminate the generation.
For simplicity, \our{} is trained with simple cross-entropy loss, without using CIDEr optimization.
During inference, we generate the caption tokens one by one in an autoregressive manner.
Table~\ref{tbl:results:vqa_nlvr2_captioning} presents the results on COCO captioning.
\our{} outperforms all previous models trained with cross-entropy loss, creating a new state-of-the-art image captioning result.
The results demonstrate the superiority of \our{} for vision-language generation.

\paragraph{Image-Text Retrieval}

\begin{table*}[t]
\centering
\small
\begin{tabular}{@{}l@{\hskip1pt} @{\hskip1pt}c@{ \hskip1pt} @{\hskip1pt}c@{ \hskip1pt} @{\hskip1pt}c@{ \hskip1pt} @{\hskip1pt}c@{ \hskip1pt} @{\hskip1pt}c@{ \hskip1pt} @{\hskip1pt}c@{ \hskip1pt} | @{ \hskip2pt}c@{ \hskip1pt} @{\hskip1pt}c@{ \hskip1pt} @{\hskip1pt}c@{ \hskip1pt} @{\hskip1pt}c@{ \hskip1pt} @{\hskip1pt}c@{ \hskip1pt} @{\hskip1pt}c@{} }
\toprule
\multirow{3}{*}{\bf Model} & \multicolumn{6}{c}{\bf MSCOCO (5K test set)} & \multicolumn{6}{c}{\bf Flickr30K (1K test set)} \\
 & \multicolumn{3}{c}{Image $\rightarrow$ Text} & \multicolumn{3}{c}{Text $\rightarrow$ Image} & \multicolumn{3}{c}{Image $\rightarrow$ Text} & \multicolumn{3}{c}{Text $\rightarrow$ Image} \\
 \cmidrule(lr){2-4} \cmidrule(lr){5-7} \cmidrule(lr){8-10} \cmidrule(lr){11-13}
 & R@1 & R@5 & R@10 & R@1 & R@5 & R@10 & R@1 & R@5 & R@10 & R@1 & R@5 & R@10 \\
\midrule
\multicolumn{13}{l}{\textit{Fusion-encoder models}} \\
UNITER~\citep{uniter} & 65.7 & 88.6 & 93.8 & 52.9 & 79.9 & 88.0 & 87.3 & 98.0 & 99.2 & 75.6 & 94.1 & 96.8 \\
VILLA~\citep{villa} & - & - & - & - & - & - & 87.9 & 97.5 & 98.8 & 76.3 & 94.2 & 96.8 \\
Oscar~\citep{oscar} & 73.5 & 92.2 & 96.0 & 57.5 & 82.8 & 89.8 & - & - & - & - & - & - \\
VinVL~\citep{vinvl} & 75.4 & 92.9 & 96.2 & 58.8 & 83.5 & 90.3 & - & - & - & - & - & - \\
\midrule
\multicolumn{13}{l}{\textit{Dual encoder + Fusion encoder reranking}} \\
ALBEF~\citep{albef} & 77.6 & 94.3 & 97.2 & 60.7 & 84.3 & 90.5 & 95.9 & 99.8 & \bf 100.0 & 85.6 & 97.5 & 98.9 \\
BLIP~\citep{blip} & 82.4 & 95.4 & 97.9 & 65.1 & 86.3 & 91.8 & 97.4 & 99.8 & 99.9 & 87.6 & 97.7 & 99.0 \\
\midrule
\multicolumn{13}{l}{\textit{Dual-encoder models}} \\
ALIGN~\citep{align} & 77.0 & 93.5 & 96.9 & 59.9 & 83.3 & 89.8 & 95.3 & 99.8 & \bf 100.0 & 84.9 & 97.4 & 98.6 \\
FILIP~\citep{filip} & 78.9 & 94.4 & 97.4 & 61.2 & 84.3 & 90.6 & 96.6 & \bf 100.0 & \bf 100.0 & 87.1 & 97.7 & 99.1 \\
Florence~\citep{florence} & 81.8 & 95.2 & - & 63.2 & 85.7 & - & 97.2 & 99.9 & - & 87.9 & 98.1 & - \\
\bf \our{} & \bf 84.8 & \bf 96.5 & \bf 98.3 & \bf 67.2 & \bf 87.7 & \bf 92.8 & \bf 98.0 & \bf 100.0 & \bf 100.0 & \bf 90.3 & \bf 98.7 & \bf 99.5 \\
\bottomrule
\end{tabular}
\caption{
Finetuning results of image-to-text retrieval and text-to-image retrieval on COCO and Flickr30K.
Notice that dual-encoder models are more efficient than fusion-encoder-based models for the retrieval tasks.
}
\label{tbl:results:finetuned_retrieval}
\end{table*}

\begin{table*}[t]
\centering
\begin{tabular}{@{}l@{\hskip4pt} @{ \hskip4pt}c@{\hskip4pt} @{\hskip4pt}c@{\hskip4pt} @{\hskip4pt}c@{\hskip4pt} @{\hskip4pt}c@{\hskip4pt} @{\hskip4pt}c@{\hskip4pt} @{\hskip4pt}c@{}}
\toprule
\multirow{3}{*}{\bf Model} & \multicolumn{6}{c}{\bf Flickr30K (1K test set)} \\
 & \multicolumn{3}{c}{Image $\rightarrow$ Text} & \multicolumn{3}{c}{Text $\rightarrow$ Image} \\
 \cmidrule(lr){2-4} \cmidrule(lr){5-7}
 & R@1 & R@5 & R@10 & R@1 & R@5 & R@10 \\
\midrule
FLAVA~\citep{flava} & 67.7 & 94.0 & - & 65.2 & 89.4 & - \\
CLIP~\citep{clip} & 88.0 & 98.7 & 99.4 & 68.7 & 90.6 & 95.2 \\
ALIGN~\citep{align} & 88.6 & 98.7 & 99.7 & 75.7 & 93.8 & 96.8 \\
FILIP~\citep{filip} & 89.8 & 99.2 & 99.8 & 75.0 & 93.4 & 96.3 \\
Florence~\citep{florence} & 90.9 & 99.1 & - & 76.7 & 93.6 & - \\
Flamingo~\citep{flamingo} & 89.3 & 98.8 & 99.7 & 79.5 & 95.3 & \bf 97.9 \\
CoCa~\citep{coca} & 92.5 & 99.5 & 99.9 & 80.4 & \bf 95.7 & 97.7 \\
\midrule
\bf \our{} & \bf 94.9 & \bf 99.9 & \bf 100.0 & \bf 81.5 & 95.6 & 97.8 \\
\bottomrule
\end{tabular}
\caption{
Zero-shot image-to-text retrieval and text-to-image retrieval on Flickr30K.
}
\label{tbl:results:zeroshot_retrieval}
\end{table*}

The task is to measure the similarity between images and texts.
There are two directions depending on the modality of the retrieved target: image-to-text retrieval, and text-to-image retrieval.
Two popular retrieval benchmarks, i.e., COCO~\citep{coco}, and Flickr30K~\citep{flickr30k}, are used to evaluate the model.
Following previous work~\citep{vinvl,vilt}, we use the Karpathy split~\citep{karpathysplit} for the two benchmarks.
\our{} is finetuned as a dual encoder for efficient image-text retrieval.
Dual-encoder models separately encode images and texts to obtain their representations. Then we calculate the cosine similarity scores of these representations.
Dual-encoder models are more efficient than fusion-encoder models. Because they do not have to jointly encode all possible image-text pairs.

We directly finetune \our{} on COCO and Flickr30K, although the model is not pretrained with image-text contrastive loss.
Surprisingly, \our{} outperforms previous state-of-the-art models only using a small amount of contrastive training.
The results demonstrate that \our{} effectively learns alignments between images and texts via masked data modeling. 
In order to improve the performance, we perform intermediate finetuning with an image-text contrastive objective on the pretraining image-text pairs.
We finetune the model with much fewer steps than pretraining.
Then we use the model to evaluate zero-shot and finetuned image-text retrieval.
The finetuned results are present in Table~\ref{tbl:results:finetuned_retrieval}, dual-encoder \our{} outperforms prior models by a large margin, achieving $3.0$/$4.0$ absolute improvement on COCO top-$1$ image-to-text/text-to-image retrieval, and $0.8$/$2.4$ absolute improvement on Flickr30K top-$1$ image-to-text/text-to-image retrieval.
\our{} also significantly outperforms fusion-encoder-based models, which require more computation cost for inference.
As present in Table~\ref{tbl:results:zeroshot_retrieval}, \our{} also achieves better performance than previous models on Flickr30K zero-shot retrieval.

\subsection{Vision Downstream Tasks}
\label{exp:vision}

In addition to vision-language downstream tasks, \our{} can be transferred to a wide range of vision downstream tasks, including object detection, instance segmentation, semantic segmentation, and image classification.
The number of effective parameters is comparable to ViT-giant~\citep{scaling:vit}, i.e., about 1B, when \our{} is used as a vision encoder.

\paragraph{Object Detection and Instance Segmentation} 

\begin{table*}[t]
\centering
\begin{tabular}{@{}lcccc@{}}
\toprule
\multirow{2}{*}{\bf Model} & \multirow{2}{*}{\bf Extra OD Data} & \multirow{2}{*}{\bf \tabincell{c}{Maximum \\ Image Size}} & \multicolumn{2}{c}{\bf COCO test-dev} \\
 & & & \boxAP{} & \maskAP{} \\
\midrule
ViT-Adapter~\citep{vit-adapter} & - & 1600 & 60.1 & 52.1 \\
DyHead~\citep{dyhead} & ImageNet-Pseudo Labels & 2000 & 60.6 & - \\
Soft Teacher~\citep{soft_teacher} & Object365 & - & 61.3 & 53.0 \\
GLIP~\citep{glip} & FourODs & - & 61.5 & - \\
GLIPv2~\citep{glipv2} & FourODs & - & 62.4 & - \\
Florence~\citep{florence} & FLOD-9M & 2500 & 62.4 & - \\
SwinV2-G~\citep{swinv2} & Object365 & 1536 & 63.1 & 54.4 \\
Mask DINO~\citep{mask_dino} & Object365 & 1280 & - & 54.7 \\
DINO~\citep{dino-od} & Object365 & 2000 & 63.3 & - \\
\midrule
\bf \our{} & Object365 & 1280 & \bf 63.7 & \bf 54.8 \\
\bottomrule
\end{tabular}
\caption{Results of object detection and instance segmentation on COCO benchmark.
\our{} uses Cascade Mask R-CNN~\citep{cascade-mask-rcnn} as the detection head.
Our results are reported with multi-scale evaluation.
We report the maximum image size used for training.
FLOD-9M and FourODs also contain Object365.
The results of the comparison systems are from the \href{https://paperswithcode.com/sota/object-detection-on-coco}{paperswithcode.com} leaderboard (timestamp: 08/22/2022).
}
\label{tbl:results:cocood}
\end{table*}

We conduct finetuning experiments on the COCO 2017 benchmark~\citep{coco}, which consists of $118$k training, $5$k validation, and $20$k test-dev images.
We use \our{} as the backbone and follow ViTDet~\citep{vitdet}, including a simple feature pyramid and window attention, for the object detection and instance segmentation tasks.
Following common practices~\citep{swinv2,dino-od}, we first conduct intermediate finetuning on the Objects365~\citep{object365} dataset. Then we finetune the model on the COCO dataset.
Soft-NMS~\citep{soft-nms} is used during inference.
Table~\ref{tbl:results:cocood} compares \our{} with previous state-of-the-art models on COCO object detection and instance segmentation.
\our{} achieves the best results on the COCO test-dev set with a smaller image size used for finetuning, reaching up to $63.7$ box AP and $54.8$ mask AP.

\paragraph{Semantic Segmentation} 

\begin{table*}[t]
\centering
\begin{tabular}{@{}lccc@{}}
\toprule
\multirow{2}{*}{\bf Model} & \multirow{2}{*}{\bf Crop Size} & \multicolumn{2}{c}{\bf ADE20K} \\
 & & mIoU & +MS \\
\midrule
HorNet~\citep{HorNet} & $640^2$ & 57.5 & 57.9 \\
SeMask~\citep{jain2021semask} & $640^2$ & 57.0 & 58.3 \\
SwinV2-G~\citep{swinv2} & $896^2$ & 59.3 & 59.9 \\
ViT-Adapter~\citep{vit-adapter} & $896^2$ & 59.4 & 60.5 \\
Mask DINO~\citep{mask_dino} & - & 59.5 & 60.8 \\
FD-SwinV2-G~\citep{fd-swin} & $896^2$ & - & 61.4 \\
\midrule
\bf \our{} & $896^2$ & \bf 62.0 & \bf 62.8 \\
\bottomrule
\end{tabular}
\caption[Caption protect]{Results of semantic segmentation on ADE20K.
``MS'' is short for multi-scale.
The results of the comparison systems are from the \href{https://paperswithcode.com/sota/semantic-segmentation-on-ade20k}{paperswithcode.com} leaderboard (timestamp: 08/22/2022).
}
\label{tbl:results:ade20k}
\end{table*}

Semantic segmentation aims to predict the label for each pixel of the given image.
We evaluate \our{} on the challenging ADE20K dataset~\citep{ade20k}, which includes $150$ semantic categories.
ADE20K contains $20$k images for training and $2$k images for validation.
We directly follow the task transfer settings of ViT-Adapter~\citep{vit-adapter}.
We use a dense prediction task adapter and employ Mask2Former~\citep{mask2former} as the segmentation framework.
As shown in Table~\ref{tbl:results:ade20k}, \our{} creates a new state-of-the-art result with $62.8$ mIoU, outperforming FD-SwinV2~\citep{fd-swin} giant model with 3B parameters by $1.4$ points.
It shows that \our{} achieves superior performance on the dense prediction task.

\paragraph{Image Classification}

\begin{table*}[t]
\centering
\begin{tabular}{@{}lccc@{}}
\toprule
\bf Model & \bf Extra Data & \bf Image Size & \bf ImageNet \\
\midrule
\multicolumn{4}{l}{~\textit{With extra \textbf{private} image-tag data}} \\
SwinV2-G~\citep{swinv2} & IN-22K-ext-70M & $640^2$ & \textcolor{lightgray}{90.2} \\
ViT-G~\citep{scaling:vit} & JFT-3B & $518^2$ & \textcolor{lightgray}{90.5} \\
CoAtNet-7~\citep{coatnet} & JFT-3B & $512^2$ & \textcolor{lightgray}{90.9} \\
Model Soups~\citep{modelsoups} & JFT-3B & $500^2$ & \textcolor{lightgray}{91.0} \\
CoCa~\citep{coca} & JFT-3B & $576^2$ & \textcolor{lightgray}{91.0} \\
\midrule
\multicolumn{4}{l}{~\textit{With only \textbf{public} image-tag data}} \\
\beit{}~\citep{beit} & IN-21K & $512^2$ & 88.6 \\
CoAtNet-4~\citep{coatnet} & IN-21K & $512^2$ & 88.6 \\
MaxViT~\citep{maxvit} & IN-21K & $512^2$ & 88.7 \\
MViTv2~\citep{mvitv2} & IN-21K & $512^2$ & 88.8 \\
FD-CLIP~\citep{fd-swin} & IN-21K & $336^2$ & 89.0 \\
\bf \our{} & IN-21K & $336^2$ & \bf 89.6 \\
\bottomrule
\end{tabular}
\caption{Top-1 accuracy on ImageNet-1K.
}
\label{tbl:results:in1k}
\end{table*}

We evaluate the model on ImageNet-1K~\citep{imagenet}, which contains $1.28$M training images and $50$k validation images in $1$k classes.
Rather than appending a task layer to the vision encoder~\citep{vit,beit}, we formulate the task as an image-to-text retrieval task.
We use the category names as texts to construct image-text pairs.
\our{} is trained as a dual encoder to find the most relevant label for an image.
During inference, we first compute the feature embeddings of possible class names and the feature embedding of the image.
Their cosine similarity scores are then calculated to predict the most probable label for each image.
Table~\ref{tbl:results:in1k} reports the results on ImageNet-1K. 
We first perform intermediate finetuning on ImageNet-21K, then we train the model on ImageNet-1K.
For a fair comparison, we compare with the previous models only using public image-tag data.
\our{} outperforms prior models, creating a new state-of-the-art result when only using public image-tag data.

\section{Conclusion}

In this paper, we present \our{}, a general-purpose multimodal foundation model, which achieves state-of-the-art performance across a wide range of vision and vision-language benchmarks.
The key idea of \our{} is that image can be modeled as a foreign language, so that we can conduct masked ``language'' modeling over images, texts, and image-text pairs in a unified way.
We also demonstrate that \multiway{} can effectively model different vision and vision-language tasks, making it an intriguing option for general-purpose modeling.
\our{} is simple and effective, and is a promising direction for scaling up multimodal foundation models.
For future work, we are working on pretraining multilingual \our{} and including more modalities (e.g., audio) in \our{} to facilitate the cross-lingual and cross-modality transfer, and advance the big convergence of large-scale pretraining across tasks, languages, and modalities.
We are also interested in enabling in-context learning capability for multimodal foundation models by combining the strength of \our{} and MetaLM~\cite{metalm}.

\newpage

\bibliographystyle{alpha}
\bibliography{beit3}

\newcommand{\etalchar}[1]{$^{#1}$}
\begin{thebibliography}{WBDW21}

\bibitem[ADL{\etalchar{+}}22]{flamingo}
Jean{-}Baptiste Alayrac, Jeff Donahue, Pauline Luc, Antoine Miech, Iain Barr,
  Yana Hasson, Karel Lenc, Arthur Mensch, Katie Millican, Malcolm Reynolds,
  Roman Ring, Eliza Rutherford, Serkan Cabi, Tengda Han, Zhitao Gong, Sina
  Samangooei, Marianne Monteiro, Jacob Menick, Sebastian Borgeaud, Andrew
  Brock, Aida Nematzadeh, Sahand Sharifzadeh, Mikolaj Binkowski, Ricardo
  Barreira, Oriol Vinyals, Andrew Zisserman, and Karen Simonyan.
\newblock Flamingo: a visual language model for few-shot learning.
\newblock {\em CoRR}, abs/2204.14198, 2022.

\bibitem[AHB{\etalchar{+}}18]{bottom_up_attn}
Peter Anderson, Xiaodong He, Chris Buehler, Damien Teney, Mark Johnson, Stephen
  Gould, and Lei Zhang.
\newblock Bottom-up and top-down attention for image captioning and visual
  question answering.
\newblock In {\em 2018 {IEEE} Conference on Computer Vision and Pattern
  Recognition, {CVPR} 2018, Salt Lake City, UT, USA, June 18-22, 2018}, pages
  6077--6086. Computer Vision Foundation / {IEEE} Computer Society, 2018.

\bibitem[BDPW22]{beit}
Hangbo Bao, Li~Dong, Songhao Piao, and Furu Wei.
\newblock {BEiT}: {BERT} pre-training of image transformers.
\newblock In {\em International Conference on Learning Representations}, 2022.

\bibitem[BDW{\etalchar{+}}20]{unilm2}
Hangbo Bao, Li~Dong, Furu Wei, Wenhui Wang, Nan Yang, Xiaodong Liu, Yu~Wang,
  Jianfeng Gao, Songhao Piao, Ming Zhou, and Hsiao{-}Wuen Hon.
\newblock {UniLMv2}: Pseudo-masked language models for unified language model
  pre-training.
\newblock In {\em Proceedings of the 37th International Conference on Machine
  Learning, {ICML} 2020, 13-18 July 2020, Virtual Event}, volume 119 of {\em
  Proceedings of Machine Learning Research}, pages 642--652. {PMLR}, 2020.

\bibitem[BDW{\etalchar{+}}21]{s2s-ft}
Hangbo Bao, Li~Dong, Wenhui Wang, Nan Yang, and Furu Wei.
\newblock s2s-ft: Fine-tuning pretrained transformer encoders for
  sequence-to-sequence learning.
\newblock {\em CoRR}, abs/2110.13640, 2021.

\bibitem[BSCD17]{soft-nms}
Navaneeth Bodla, Bharat Singh, Rama Chellappa, and Larry~S. Davis.
\newblock Soft-nms - improving object detection with one line of code.
\newblock In {\em {IEEE} International Conference on Computer Vision, {ICCV}
  2017, Venice, Italy, October 22-29, 2017}, pages 5562--5570. {IEEE} Computer
  Society, 2017.

\bibitem[BWDW22]{vlbeit}
Hangbo Bao, Wenhui Wang, Li~Dong, and Furu Wei.
\newblock {VL-BEiT}: Generative vision-language pretraining.
\newblock {\em ArXiv}, abs/2206.01127, 2022.

\bibitem[CDW{\etalchar{+}}22]{vit-adapter}
Zhe Chen, Yuchen Duan, Wenhai Wang, Junjun He, Tong Lu, Jifeng Dai, and
  Yu~Qiao.
\newblock Vision transformer adapter for dense predictions.
\newblock {\em CoRR}, abs/2205.08534, 2022.

\bibitem[CLY{\etalchar{+}}20]{uniter}
Yen{-}Chun Chen, Linjie Li, Licheng Yu, Ahmed~El Kholy, Faisal Ahmed, Zhe Gan,
  Yu~Cheng, and Jingjing Liu.
\newblock {UNITER:} universal image-text representation learning.
\newblock In Andrea Vedaldi, Horst Bischof, Thomas Brox, and Jan{-}Michael
  Frahm, editors, {\em Computer Vision - {ECCV} 2020 - 16th European
  Conference, Glasgow, UK, August 23-28, 2020, Proceedings, Part {XXX}}, volume
  12375 of {\em Lecture Notes in Computer Science}, pages 104--120. Springer,
  2020.

\bibitem[CMS{\etalchar{+}}21]{mask2former}
Bowen Cheng, Ishan Misra, Alexander~G. Schwing, Alexander Kirillov, and Rohit
  Girdhar.
\newblock Masked-attention mask transformer for universal image segmentation.
\newblock {\em CoRR}, abs/2112.01527, 2021.

\bibitem[CSDS21]{cc12m}
Soravit Changpinyo, Piyush Sharma, Nan Ding, and Radu Soricut.
\newblock Conceptual 12m: Pushing web-scale image-text pre-training to
  recognize long-tail visual concepts.
\newblock In {\em {IEEE} Conference on Computer Vision and Pattern Recognition,
  {CVPR} 2021, virtual, June 19-25, 2021}, pages 3558--3568. Computer Vision
  Foundation / {IEEE}, 2021.

\bibitem[CV21]{cascade-mask-rcnn}
Zhaowei Cai and Nuno Vasconcelos.
\newblock Cascade {R-CNN:} high quality object detection and instance
  segmentation.
\newblock {\em {IEEE} Trans. Pattern Anal. Mach. Intell.}, 43(5):1483--1498,
  2021.

\bibitem[DBK{\etalchar{+}}20]{vit}
Alexey Dosovitskiy, Lucas Beyer, Alexander Kolesnikov, Dirk Weissenborn,
  Xiaohua Zhai, Thomas Unterthiner, Mostafa Dehghani, Matthias Minderer, Georg
  Heigold, Sylvain Gelly, et~al.
\newblock An image is worth 16x16 words: Transformers for image recognition at
  scale.
\newblock {\em preprint arXiv:2010.11929}, 2020.

\bibitem[DCLT19]{bert}
Jacob Devlin, Ming{-}Wei Chang, Kenton Lee, and Kristina Toutanova.
\newblock {BERT:} pre-training of deep bidirectional transformers for language
  understanding.
\newblock In Jill Burstein, Christy Doran, and Thamar Solorio, editors, {\em
  Proceedings of the 2019 Conference of the North American Chapter of the
  Association for Computational Linguistics: Human Language Technologies,
  {NAACL-HLT} 2019, Minneapolis, MN, USA, June 2-7, 2019, Volume 1 (Long and
  Short Papers)}, pages 4171--4186. Association for Computational Linguistics,
  2019.

\bibitem[DCX{\etalchar{+}}21]{dyhead}
Xiyang Dai, Yinpeng Chen, Bin Xiao, Dongdong Chen, Mengchen Liu, Lu~Yuan, and
  Lei Zhang.
\newblock Dynamic head: Unifying object detection heads with attentions.
\newblock In {\em {IEEE} Conference on Computer Vision and Pattern Recognition,
  {CVPR} 2021, virtual, June 19-25, 2021}, pages 7373--7382. Computer Vision
  Foundation / {IEEE}, 2021.

\bibitem[DLLT21]{coatnet}
Zihang Dai, Hanxiao Liu, Quoc~V. Le, and Mingxing Tan.
\newblock Coatnet: Marrying convolution and attention for all data sizes.
\newblock In Marc'Aurelio Ranzato, Alina Beygelzimer, Yann~N. Dauphin, Percy
  Liang, and Jennifer~Wortman Vaughan, editors, {\em Advances in Neural
  Information Processing Systems 34: Annual Conference on Neural Information
  Processing Systems 2021, NeurIPS 2021, December 6-14, 2021, virtual}, pages
  3965--3977, 2021.

\bibitem[DYW{\etalchar{+}}19]{unilm}
Li~Dong, Nan Yang, Wenhui Wang, Furu Wei, Xiaodong Liu, Yu~Wang, Jianfeng Gao,
  Ming Zhou, and Hsiao{-}Wuen Hon.
\newblock Unified language model pre-training for natural language
  understanding and generation.
\newblock In {\em Advances in Neural Information Processing Systems 32: Annual
  Conference on Neural Information Processing Systems 2019, NeurIPS 2019,
  December 8-14, 2019, Vancouver, BC, Canada}, pages 13042--13054, 2019.

\bibitem[GCL{\etalchar{+}}20]{villa}
Zhe Gan, Yen{-}Chun Chen, Linjie Li, Chen Zhu, Yu~Cheng, and Jingjing Liu.
\newblock Large-scale adversarial training for vision-and-language
  representation learning.
\newblock In Hugo Larochelle, Marc'Aurelio Ranzato, Raia Hadsell,
  Maria{-}Florina Balcan, and Hsuan{-}Tien Lin, editors, {\em Advances in
  Neural Information Processing Systems 33: Annual Conference on Neural
  Information Processing Systems 2020, NeurIPS 2020, December 6-12, 2020,
  virtual}, 2020.

\bibitem[GKS{\etalchar{+}}17]{vqa}
Yash Goyal, Tejas Khot, Douglas Summers{-}Stay, Dhruv Batra, and Devi Parikh.
\newblock Making the {V} in {VQA} matter: Elevating the role of image
  understanding in visual question answering.
\newblock In {\em 2017 {IEEE} Conference on Computer Vision and Pattern
  Recognition, {CVPR} 2017, Honolulu, HI, USA, July 21-26, 2017}, pages
  6325--6334. {IEEE} Computer Society, 2017.

\bibitem[HSD{\etalchar{+}}22]{metalm}
Yaru Hao, Haoyu Song, Li~Dong, Shaohan Huang, Zewen Chi, Wenhui Wang, Shuming
  Ma, and Furu Wei.
\newblock Language models are general-purpose interfaces.
\newblock {\em ArXiv}, abs/2206.06336, 2022.

\bibitem[HSL{\etalchar{+}}16]{drop_path}
Gao Huang, Yu~Sun, Zhuang Liu, Daniel Sedra, and Kilian~Q. Weinberger.
\newblock Deep networks with stochastic depth.
\newblock In Bastian Leibe, Jiri Matas, Nicu Sebe, and Max Welling, editors,
  {\em Computer Vision - {ECCV} 2016 - 14th European Conference, Amsterdam, The
  Netherlands, October 11-14, 2016, Proceedings, Part {IV}}, volume 9908 of
  {\em Lecture Notes in Computer Science}, pages 646--661. Springer, 2016.

\bibitem[JSO{\etalchar{+}}21]{jain2021semask}
Jitesh Jain, Anukriti Singh, Nikita Orlov, Zilong Huang, Jiachen Li, Steven
  Walton, and Humphrey Shi.
\newblock Semask: Semantically masking transformer backbones for effective
  semantic segmentation.
\newblock {\em arXiv}, 2021.

\bibitem[JYX{\etalchar{+}}21]{align}
Chao Jia, Yinfei Yang, Ye~Xia, Yi{-}Ting Chen, Zarana Parekh, Hieu Pham,
  Quoc~V. Le, Yun{-}Hsuan Sung, Zhen Li, and Tom Duerig.
\newblock Scaling up visual and vision-language representation learning with
  noisy text supervision.
\newblock In Marina Meila and Tong Zhang, editors, {\em Proceedings of the 38th
  International Conference on Machine Learning, {ICML} 2021, 18-24 July 2021,
  Virtual Event}, volume 139 of {\em Proceedings of Machine Learning Research},
  pages 4904--4916. {PMLR}, 2021.

\bibitem[KF15]{karpathysplit}
Andrej Karpathy and Li~Fei{-}Fei.
\newblock Deep visual-semantic alignments for generating image descriptions.
\newblock In {\em {IEEE} Conference on Computer Vision and Pattern Recognition,
  {CVPR} 2015, Boston, MA, USA, June 7-12, 2015}, pages 3128--3137. {IEEE}
  Computer Society, 2015.

\bibitem[KR18]{sentencepiece}
Taku Kudo and John Richardson.
\newblock {S}entence{P}iece: A simple and language independent subword
  tokenizer and detokenizer for neural text processing.
\newblock In {\em Proceedings of the 2018 Conference on Empirical Methods in
  Natural Language Processing: System Demonstrations}, pages 66--71, Brussels,
  Belgium, November 2018. Association for Computational Linguistics.

\bibitem[KSK21]{vilt}
Wonjae Kim, Bokyung Son, and Ildoo Kim.
\newblock {ViLT}: Vision-and-language transformer without convolution or region
  supervision.
\newblock In Marina Meila and Tong Zhang, editors, {\em Proceedings of the 38th
  International Conference on Machine Learning, {ICML} 2021, 18-24 July 2021,
  Virtual Event}, volume 139 of {\em Proceedings of Machine Learning Research},
  pages 5583--5594. {PMLR}, 2021.

\bibitem[KZG{\etalchar{+}}17]{vg}
Ranjay Krishna, Yuke Zhu, Oliver Groth, Justin Johnson, Kenji Hata, Joshua
  Kravitz, Stephanie Chen, Yannis Kalantidis, Li{-}Jia Li, David~A. Shamma,
  Michael~S. Bernstein, and Li~Fei{-}Fei.
\newblock Visual genome: Connecting language and vision using crowdsourced
  dense image annotations.
\newblock {\em Int. J. Comput. Vis.}, 123(1):32--73, 2017.

\bibitem[LH19]{adamw}
Ilya Loshchilov and Frank Hutter.
\newblock Decoupled weight decay regularization.
\newblock In {\em 7th International Conference on Learning Representations,
  {ICLR} 2019, New Orleans, LA, USA, May 6-9, 2019}. OpenReview.net, 2019.

\bibitem[LHL{\etalchar{+}}21]{swinv2}
Ze~Liu, Han Hu, Yutong Lin, Zhuliang Yao, Zhenda Xie, Yixuan Wei, Jia Ning, Yue
  Cao, Zheng Zhang, Li~Dong, Furu Wei, and Baining Guo.
\newblock Swin transformer {V2:} scaling up capacity and resolution.
\newblock {\em CoRR}, abs/2111.09883, 2021.

\bibitem[LLXH22]{blip}
Junnan Li, Dongxu Li, Caiming Xiong, and Steven C.~H. Hoi.
\newblock {BLIP:} bootstrapping language-image pre-training for unified
  vision-language understanding and generation.
\newblock In Kamalika Chaudhuri, Stefanie Jegelka, Le~Song, Csaba
  Szepesv{\'{a}}ri, Gang Niu, and Sivan Sabato, editors, {\em International
  Conference on Machine Learning, {ICML} 2022, 17-23 July 2022, Baltimore,
  Maryland, {USA}}, volume 162 of {\em Proceedings of Machine Learning
  Research}, pages 12888--12900. {PMLR}, 2022.

\bibitem[LMB{\etalchar{+}}14]{coco}
Tsung{-}Yi Lin, Michael Maire, Serge~J. Belongie, James Hays, Pietro Perona,
  Deva Ramanan, Piotr Doll{\'{a}}r, and C.~Lawrence Zitnick.
\newblock Microsoft {COCO:} common objects in context.
\newblock In David~J. Fleet, Tom{\'{a}}s Pajdla, Bernt Schiele, and Tinne
  Tuytelaars, editors, {\em Computer Vision - {ECCV} 2014 - 13th European
  Conference, Zurich, Switzerland, September 6-12, 2014, Proceedings, Part
  {V}}, volume 8693 of {\em Lecture Notes in Computer Science}, pages 740--755.
  Springer, 2014.

\bibitem[LMGH22]{vitdet}
Yanghao Li, Hanzi Mao, Ross~B. Girshick, and Kaiming He.
\newblock Exploring plain vision transformer backbones for object detection.
\newblock {\em CoRR}, abs/2203.16527, 2022.

\bibitem[LOG{\etalchar{+}}19]{roberta}
Yinhan Liu, Myle Ott, Naman Goyal, Jingfei Du, Mandar Joshi, Danqi Chen, Omer
  Levy, Mike Lewis, Luke Zettlemoyer, and Veselin Stoyanov.
\newblock Roberta: {A} robustly optimized {BERT} pretraining approach.
\newblock {\em CoRR}, abs/1907.11692, 2019.

\bibitem[LSG{\etalchar{+}}21]{albef}
Junnan Li, Ramprasaath~R. Selvaraju, Akhilesh~Deepak Gotmare, Shafiq~R. Joty,
  Caiming Xiong, and Steven C.~H. Hoi.
\newblock Align before fuse: Vision and language representation learning with
  momentum distillation.
\newblock {\em CoRR}, abs/2107.07651, 2021.

\bibitem[LWF{\etalchar{+}}22]{mvitv2}
Yanghao Li, Chao-Yuan Wu, Haoqi Fan, Karttikeya Mangalam, Bo~Xiong, Jitendra
  Malik, and Christoph Feichtenhofer.
\newblock Mvitv2: Improved multiscale vision transformers for classification
  and detection.
\newblock In {\em Proceedings of the IEEE/CVF Conference on Computer Vision and
  Pattern Recognition}, pages 4804--4814, 2022.

\bibitem[LYL{\etalchar{+}}20]{oscar}
Xiujun Li, Xi~Yin, Chunyuan Li, Pengchuan Zhang, Xiaowei Hu, Lei Zhang, Lijuan
  Wang, Houdong Hu, Li~Dong, Furu Wei, Yejin Choi, and Jianfeng Gao.
\newblock Oscar: Object-semantics aligned pre-training for vision-language
  tasks.
\newblock In Andrea Vedaldi, Horst Bischof, Thomas Brox, and Jan{-}Michael
  Frahm, editors, {\em Computer Vision - {ECCV} 2020 - 16th European
  Conference, Glasgow, UK, August 23-28, 2020, Proceedings, Part {XXX}}, volume
  12375 of {\em Lecture Notes in Computer Science}, pages 121--137. Springer,
  2020.

\bibitem[LZX{\etalchar{+}}22]{mask_dino}
Feng Li, Hao Zhang, Huaizhe Xu, Shilong Liu, Lei Zhang, Lionel~M. Ni, and
  Heung{-}Yeung Shum.
\newblock Mask {DINO:} towards {A} unified transformer-based framework for
  object detection and segmentation.
\newblock {\em CoRR}, abs/2206.02777, 2022.

\bibitem[LZZ{\etalchar{+}}21]{glip}
Liunian~Harold Li, Pengchuan Zhang, Haotian Zhang, Jianwei Yang, Chunyuan Li,
  Yiwu Zhong, Lijuan Wang, Lu~Yuan, Lei Zhang, Jenq{-}Neng Hwang, Kai{-}Wei
  Chang, and Jianfeng Gao.
\newblock Grounded language-image pre-training.
\newblock {\em CoRR}, abs/2112.03857, 2021.

\bibitem[OKB11]{sbu}
Vicente Ordonez, Girish Kulkarni, and Tamara~L. Berg.
\newblock Im2text: Describing images using 1 million captioned photographs.
\newblock In John Shawe{-}Taylor, Richard~S. Zemel, Peter~L. Bartlett, Fernando
  C.~N. Pereira, and Kilian~Q. Weinberger, editors, {\em Advances in Neural
  Information Processing Systems 24: 25th Annual Conference on Neural
  Information Processing Systems 2011. Proceedings of a meeting held 12-14
  December 2011, Granada, Spain}, pages 1143--1151, 2011.

\bibitem[PDB{\etalchar{+}}22]{beitv2}
Zhiliang Peng, Li~Dong, Hangbo Bao, Qixiang Ye, and Furu Wei.
\newblock Beit v2: Masked image modeling with vector-quantized visual
  tokenizers.
\newblock {\em CoRR}, abs/2208.06366, 2022.

\bibitem[PWC{\etalchar{+}}15]{flickr30k}
Bryan~A. Plummer, Liwei Wang, Chris~M. Cervantes, Juan~C. Caicedo, Julia
  Hockenmaier, and Svetlana Lazebnik.
\newblock Flickr30k entities: Collecting region-to-phrase correspondences for
  richer image-to-sentence models.
\newblock In {\em 2015 {IEEE} International Conference on Computer Vision,
  {ICCV} 2015, Santiago, Chile, December 7-13, 2015}, pages 2641--2649. {IEEE}
  Computer Society, 2015.

\bibitem[RDS{\etalchar{+}}15]{imagenet}
Olga Russakovsky, Jia Deng, Hao Su, Jonathan Krause, Sanjeev Satheesh, Sean Ma,
  Zhiheng Huang, Andrej Karpathy, Aditya Khosla, Michael Bernstein, Alexander~C
  Berg, and Li~Fei-Fei.
\newblock Imagenet large scale visual recognition challenge.
\newblock {\em IJCV}, 2015.

\bibitem[RKH{\etalchar{+}}21]{clip}
Alec Radford, Jong~Wook Kim, Chris Hallacy, Aditya Ramesh, Gabriel Goh,
  Sandhini Agarwal, Girish Sastry, Amanda Askell, Pamela Mishkin, Jack Clark,
  Gretchen Krueger, and Ilya Sutskever.
\newblock Learning transferable visual models from natural language
  supervision.
\newblock In Marina Meila and Tong Zhang, editors, {\em Proceedings of the 38th
  International Conference on Machine Learning, {ICML} 2021, 18-24 July 2021,
  Virtual Event}, volume 139 of {\em Proceedings of Machine Learning Research},
  pages 8748--8763. {PMLR}, 2021.

\bibitem[RNSS18]{gpt}
Alec Radford, Karthik Narasimhan, Tim Salimans, and Ilya Sutskever.
\newblock Improving language understanding by generative pre-training.
\newblock 2018.

\bibitem[RZT{\etalchar{+}}22]{HorNet}
Yongming Rao, Wenliang Zhao, Yansong Tang, Jie Zhou, Ser~Nam Lim, and Jiwen Lu.
\newblock {HorNet}: Efficient high-order spatial interactions with recursive
  gated convolutions.
\newblock {\em ArXiv}, abs/2207.14284, 2022.

\bibitem[SDGS18]{gcc}
Piyush Sharma, Nan Ding, Sebastian Goodman, and Radu Soricut.
\newblock Conceptual captions: {A} cleaned, hypernymed, image alt-text dataset
  for automatic image captioning.
\newblock In Iryna Gurevych and Yusuke Miyao, editors, {\em Proceedings of the
  56th Annual Meeting of the Association for Computational Linguistics, {ACL}
  2018, Melbourne, Australia, July 15-20, 2018, Volume 1: Long Papers}, pages
  2556--2565. Association for Computational Linguistics, 2018.

\bibitem[SHG{\etalchar{+}}21]{flava}
Amanpreet Singh, Ronghang Hu, Vedanuj Goswami, Guillaume Couairon, Wojciech
  Galuba, Marcus Rohrbach, and Douwe Kiela.
\newblock {FLAVA:} {A} foundational language and vision alignment model.
\newblock {\em CoRR}, abs/2112.04482, 2021.

\bibitem[SLZ{\etalchar{+}}19]{object365}
Shuai Shao, Zeming Li, Tianyuan Zhang, Chao Peng, Gang Yu, Xiangyu Zhang, Jing
  Li, and Jian Sun.
\newblock Objects365: {A} large-scale, high-quality dataset for object
  detection.
\newblock In {\em 2019 {IEEE/CVF} International Conference on Computer Vision,
  {ICCV} 2019, Seoul, Korea (South), October 27 - November 2, 2019}, pages
  8429--8438. {IEEE}, 2019.

\bibitem[SZZ{\etalchar{+}}19]{nlvr2}
Alane Suhr, Stephanie Zhou, Ally Zhang, Iris Zhang, Huajun Bai, and Yoav Artzi.
\newblock A corpus for reasoning about natural language grounded in
  photographs.
\newblock In Anna Korhonen, David~R. Traum, and Llu{\'{\i}}s M{\`{a}}rquez,
  editors, {\em Proceedings of the 57th Conference of the Association for
  Computational Linguistics, {ACL} 2019, Florence, Italy, July 28- August 2,
  2019, Volume 1: Long Papers}, pages 6418--6428. Association for Computational
  Linguistics, 2019.

\bibitem[TL18]{stories_data}
Trieu~H. Trinh and Quoc~V. Le.
\newblock A simple method for commonsense reasoning.
\newblock {\em ArXiv}, abs/1806.02847, 2018.

\bibitem[TTZ{\etalchar{+}}22]{maxvit}
Zhengzhong Tu, Hossein Talebi, Han Zhang, Feng Yang, Peyman Milanfar, Alan
  Bovik, and Yinxiao Li.
\newblock Maxvit: Multi-axis vision transformer.
\newblock {\em CoRR}, abs/2204.01697, 2022.

\bibitem[VSP{\etalchar{+}}17]{transformer}
Ashish Vaswani, Noam Shazeer, Niki Parmar, Jakob Uszkoreit, Llion Jones,
  Aidan~N. Gomez, Lukasz Kaiser, and Illia Polosukhin.
\newblock Attention is all you need.
\newblock In Isabelle Guyon, Ulrike von Luxburg, Samy Bengio, Hanna~M. Wallach,
  Rob Fergus, S.~V.~N. Vishwanathan, and Roman Garnett, editors, {\em Advances
  in Neural Information Processing Systems 30: Annual Conference on Neural
  Information Processing Systems 2017, December 4-9, 2017, Long Beach, CA,
  {USA}}, pages 5998--6008, 2017.

\bibitem[WBDW21]{vlmo}
Wenhui Wang, Hangbo Bao, Li~Dong, and Furu Wei.
\newblock {VLMo}: Unified vision-language pre-training with
  mixture-of-modality-experts.
\newblock {\em CoRR}, abs/2111.02358, 2021.

\bibitem[WHX{\etalchar{+}}22]{fd-swin}
Yixuan Wei, Han Hu, Zhenda Xie, Zheng Zhang, Yue Cao, Jianmin Bao, Dong Chen,
  and Baining Guo.
\newblock Contrastive learning rivals masked image modeling in fine-tuning via
  feature distillation.
\newblock {\em CoRR}, abs/2205.14141, 2022.

\bibitem[WIG{\etalchar{+}}22]{modelsoups}
Mitchell Wortsman, Gabriel Ilharco, Samir~Ya Gadre, Rebecca Roelofs,
  Raphael~Gontijo Lopes, Ari~S. Morcos, Hongseok Namkoong, Ali Farhadi, Yair
  Carmon, Simon Kornblith, and Ludwig Schmidt.
\newblock Model soups: averaging weights of multiple fine-tuned models improves
  accuracy without increasing inference time.
\newblock In Kamalika Chaudhuri, Stefanie Jegelka, Le~Song, Csaba
  Szepesv{\'{a}}ri, Gang Niu, and Sivan Sabato, editors, {\em International
  Conference on Machine Learning, {ICML} 2022, 17-23 July 2022, Baltimore,
  Maryland, {USA}}, volume 162 of {\em Proceedings of Machine Learning
  Research}, pages 23965--23998. {PMLR}, 2022.

\bibitem[WXYL18]{coloraug}
Zhirong Wu, Yuanjun Xiong, Stella~X. Yu, and Dahua Lin.
\newblock Unsupervised feature learning via non-parametric instance
  discrimination.
\newblock In {\em 2018 {IEEE} Conference on Computer Vision and Pattern
  Recognition, {CVPR} 2018, Salt Lake City, UT, USA, June 18-22, 2018}, pages
  3733--3742. Computer Vision Foundation / {IEEE} Computer Society, 2018.

\bibitem[WYM{\etalchar{+}}22]{ofa}
Peng Wang, An~Yang, Rui Men, Junyang Lin, Shuai Bai, Zhikang Li, Jianxin Ma,
  Chang Zhou, Jingren Zhou, and Hongxia Yang.
\newblock Unifying architectures, tasks, and modalities through a simple
  sequence-to-sequence learning framework.
\newblock {\em CoRR}, abs/2202.03052, 2022.

\bibitem[WYY{\etalchar{+}}21]{simvlm}
Zirui Wang, Jiahui Yu, Adams~Wei Yu, Zihang Dai, Yulia Tsvetkov, and Yuan Cao.
\newblock {SimVLM}: Simple visual language model pretraining with weak
  supervision.
\newblock {\em CoRR}, abs/2108.10904, 2021.

\bibitem[XZH{\etalchar{+}}21]{soft_teacher}
Mengde Xu, Zheng Zhang, Han Hu, Jianfeng Wang, Lijuan Wang, Fangyun Wei, Xiang
  Bai, and Zicheng Liu.
\newblock End-to-end semi-supervised object detection with soft teacher.
\newblock In {\em 2021 {IEEE/CVF} International Conference on Computer Vision,
  {ICCV} 2021, Montreal, QC, Canada, October 10-17, 2021}, pages 3040--3049.
  {IEEE}, 2021.

\bibitem[YCC{\etalchar{+}}21]{florence}
Lu~Yuan, Dongdong Chen, Yi{-}Ling Chen, Noel Codella, Xiyang Dai, Jianfeng Gao,
  Houdong Hu, Xuedong Huang, Boxin Li, Chunyuan Li, Ce~Liu, Mengchen Liu,
  Zicheng Liu, Yumao Lu, Yu~Shi, Lijuan Wang, Jianfeng Wang, Bin Xiao, Zhen
  Xiao, Jianwei Yang, Michael Zeng, Luowei Zhou, and Pengchuan Zhang.
\newblock Florence: {A} new foundation model for computer vision.
\newblock {\em CoRR}, abs/2111.11432, 2021.

\bibitem[YHH{\etalchar{+}}21]{filip}
Lewei Yao, Runhui Huang, Lu~Hou, Guansong Lu, Minzhe Niu, Hang Xu, Xiaodan
  Liang, Zhenguo Li, Xin Jiang, and Chunjing Xu.
\newblock {FILIP:} fine-grained interactive language-image pre-training.
\newblock {\em CoRR}, abs/2111.07783, 2021.

\bibitem[YWV{\etalchar{+}}22]{coca}
Jiahui Yu, Zirui Wang, Vijay Vasudevan, Legg Yeung, Mojtaba Seyedhosseini, and
  Yonghui Wu.
\newblock Coca: Contrastive captioners are image-text foundation models.
\newblock {\em CoRR}, abs/2205.01917, 2022.

\bibitem[ZKHB21]{scaling:vit}
Xiaohua Zhai, Alexander Kolesnikov, Neil Houlsby, and Lucas Beyer.
\newblock Scaling vision transformers.
\newblock {\em arXiv preprint arXiv:2106.04560}, 2021.

\bibitem[ZKZ{\etalchar{+}}15]{bookcorpus}
Yukun Zhu, Ryan Kiros, Rich Zemel, Ruslan Salakhutdinov, Raquel Urtasun,
  Antonio Torralba, and Sanja Fidler.
\newblock Aligning books and movies: Towards story-like visual explanations by
  watching movies and reading books.
\newblock In {\em Proceedings of the IEEE international conference on computer
  vision}, pages 19--27, 2015.

\bibitem[ZLH{\etalchar{+}}21]{vinvl}
Pengchuan Zhang, Xiujun Li, Xiaowei Hu, Jianwei Yang, Lei Zhang, Lijuan Wang,
  Yejin Choi, and Jianfeng Gao.
\newblock {VinVL}: Revisiting visual representations in vision-language models.
\newblock In {\em {IEEE} Conference on Computer Vision and Pattern Recognition,
  {CVPR} 2021, virtual, June 19-25, 2021}, pages 5579--5588. Computer Vision
  Foundation / {IEEE}, 2021.

\bibitem[ZLL{\etalchar{+}}22]{dino-od}
Hao Zhang, Feng Li, Shilong Liu, Lei Zhang, Hang Su, Jun Zhu, Lionel~M. Ni, and
  Heung{-}Yeung Shum.
\newblock {DINO:} {DETR} with improved denoising anchor boxes for end-to-end
  object detection.
\newblock {\em CoRR}, abs/2203.03605, 2022.

\bibitem[ZZH{\etalchar{+}}22]{glipv2}
Haotian Zhang, Pengchuan Zhang, Xiaowei Hu, Yen{-}Chun Chen, Liunian~Harold Li,
  Xiyang Dai, Lijuan Wang, Lu~Yuan, Jenq{-}Neng Hwang, and Jianfeng Gao.
\newblock Glipv2: Unifying localization and vision-language understanding.
\newblock {\em CoRR}, abs/2206.05836, 2022.

\bibitem[ZZP{\etalchar{+}}19]{ade20k}
Bolei Zhou, Hang Zhao, Xavier Puig, Tete Xiao, Sanja Fidler, Adela Barriuso,
  and Antonio Torralba.
\newblock Semantic understanding of scenes through the {ADE20K} dataset.
\newblock {\em Int. J. Comput. Vis.}, 127(3):302--321, 2019.

\end{thebibliography}

\newpage
\appendix

\section{Effects of Intermediate Finetuning for Retrieval}

As shown in Table~\ref{tbl:results:finetuned_retrieval_woitcct}, we directly finetune \our{} on COCO and Flickr30K.
\our{} still outperforms previous state-of-the-art models, even without using image-text contrastive objective during pretraining.
The results demonstrate the effectiveness of masked data modeling for learning cross-modal representations.
Next, we perform intermediate finetuning on the pretraining image-text pairs for $5$ epochs with a $16$k batch size.
The peak learning is 3e-5, with linear warmup over the first epoch.
The image input size is $224 \times 224$.
The weight decay is set to $0.05$.
We disable dropout as in pretraining and use drop path with a rate of $0.3$.
The layer-wise learning rate decay is $0.95$.
We use the AdamW~\citep{adamw} optimizer with $\beta_1=0.9$, $\beta_2=0.999$.

\begin{table*}[h]
\centering
\small
\begin{tabular}{@{}l@{\hskip1pt} @{\hskip1pt}c@{ \hskip1pt} @{\hskip1pt}c@{ \hskip1pt} @{\hskip1pt}c@{ \hskip1pt} @{\hskip1pt}c@{ \hskip1pt} @{\hskip1pt}c@{ \hskip1pt} @{\hskip1pt}c@{ \hskip1pt} | @{ \hskip2pt}c@{ \hskip1pt} @{\hskip1pt}c@{ \hskip1pt} @{\hskip1pt}c@{ \hskip1pt} @{\hskip1pt}c@{ \hskip1pt} @{\hskip1pt}c@{ \hskip1pt} @{\hskip1pt}c@{} }
\toprule
\multirow{3}{*}{\bf Model} & \multicolumn{6}{c}{\bf MSCOCO (5K test set)} & \multicolumn{6}{c}{\bf Flickr30K (1K test set)} \\
 & \multicolumn{3}{c}{Image $\rightarrow$ Text} & \multicolumn{3}{c}{Text $\rightarrow$ Image} & \multicolumn{3}{c}{Image $\rightarrow$ Text} & \multicolumn{3}{c}{Text $\rightarrow$ Image} \\
 \cmidrule(lr){2-4} \cmidrule(lr){5-7} \cmidrule(lr){8-10} \cmidrule(lr){11-13}
 & R@1 & R@5 & R@10 & R@1 & R@5 & R@10 & R@1 & R@5 & R@10 & R@1 & R@5 & R@10 \\
\midrule
\our{} & 82.7 & 96.0 & 98.2 & 65.1 & 86.6 & 92.3 & 97.5 & 99.9 & 100.0 & 89.1 & 98.6 & 99.3 \\
~~+ Intermediate Finetuning & \bf 84.8 & \bf 96.5 & \bf 98.3 & \bf 67.2 & \bf 87.7 & \bf 92.8 & \bf 98.0 & \bf 100.0 & \bf 100.0 & \bf 90.3 & \bf 98.7 & \bf 99.5 \\
\bottomrule
\end{tabular}
\caption{Finetuning results of image-text retrieval on COCO and Flickr30K.
\our{} is directly finetuned on downstream benchmarks without intermediate finetuning on the pretraining data.
}
\label{tbl:results:finetuned_retrieval_woitcct}
\end{table*}

\section{Hyperparameters Used for Pretraining}

\begin{table}[H]
\centering
\small
\begin{tabular}{l|c}
\toprule
\bf Hyperparameters & \bf \our{} \\
\midrule
Layers & 40 \\
Hidden size & 1408 \\
FFN inner hidden size & 6144 \\
Attention heads & 16 \\
Patch size & $14 \times 14$ \\
Relative positional embeddings & \xmark \\
\midrule
Training steps & 1M \\
Batch size & 6144 \\
AdamW $\epsilon$ & 1e-6 \\
AdamW $\beta$ & (0.9, 0.98) \\
Peak learning rate & 1e-3 \\
Learning rate schedule & Cosine \\
Warmup steps & 10k \\
\midrule
Gradient clipping & 3.0 \\
Dropout & \xmark \\
Drop path & 0.1 \\
Weight decay & 0.05 \\
\midrule
Data Augment & RandomResizeAndCrop \\
Input resolution & $224^2$ \\
Color jitter & 0.4 \\
\bottomrule
\end{tabular}
\vspace{2mm}
\caption{
Hyperparameters for pretraining \our{}.
}
\label{tbl:pretrain:hyperparams}
\end{table}

\section{Hyperparameters Used for Finetuning}

\begin{table}[H]
\centering
\small
\begin{tabular}{l|cc}
\toprule
\bf Hyperparameters & \bf NLVR2 & \bf VQAv2 \\
\midrule
Peak learning rate & 1e-3 & 1e-5 \\
Fine-tuning epochs & 20  & 10 \\
Warmup epochs & 5 & 1 \\
Layer-wise learning rate decay & 0.8 & 1.0 \\
Batch size & 256 & 128 \\
AdamW $\epsilon$ & \multicolumn{2}{c}{1e-8}  \\
AdamW $\beta$ & \multicolumn{2}{c}{(0.9, 0.999)} \\
Weight decay & 0.05 & 0.01 \\
Drop path & \multicolumn{2}{c}{0.4} \\
Dropout & \multicolumn{2}{c}{\xmark} \\
Input resolution & $224^2$ & $756^2$ \\
\bottomrule
\end{tabular}
\vspace{2mm}
\caption{
Hyperparameters for fine-tuning \our{} on NLVR2 and VQAv2.
}
\label{tbl:ft:vqa_nlvr2:hyperparams}
\end{table}

\begin{table}[H]
\centering
\small
\begin{tabular}{l|c}
\toprule
\bf Hyperparameters & \bf COCO Captioning \\
\midrule
Peak learning rate & 8e-6 \\
Fine-tuning steps & $16$k \\
Warmup steps & 1600 \\
Layer-wise learning rate decay & 1.0 \\
Batch size & 256 \\
AdamW $\epsilon$ & 1e-8  \\
AdamW $\beta$ & (0.9, 0.999) \\
Weight decay & 0.01 \\
Drop path & 0.3 \\
Dropout & \xmark \\
Input resolution & $392^2$ \\
Mask prob & 0.6 \\
Label smoothing $\varepsilon$ & 0.1 \\
Beam size & 3 \\
\bottomrule
\end{tabular}
\vspace{2mm}
\caption{
Hyperparameters for fine-tuning \our{} on COCO captioning.
}
\label{tbl:ft:captioning:hyperparams}
\end{table}

\begin{table}[H]
\centering
\small
\begin{tabular}{l|cc}
\toprule
\bf Hyperparameters & \bf COCO & \bf Flickr30K \\
\midrule
Peak learning rate & \multicolumn{2}{c}{1e-5} \\
Fine-tuning epochs & 15 & 20 \\
Warmup epochs & 3 & 5 \\
Layer-wise learning rate decay & \multicolumn{2}{c}{0.95} \\
Batch size & \multicolumn{2}{c}{$3$k} \\
AdamW $\epsilon$ & \multicolumn{2}{c}{1e-8}  \\
AdamW $\beta$ & \multicolumn{2}{c}{(0.9, 0.999)} \\
Weight decay & \multicolumn{2}{c}{0.05} \\
Drop path & \multicolumn{2}{c}{0.3} \\
Dropout & \multicolumn{2}{c}{\xmark} \\
Input resolution & \multicolumn{2}{c}{$420^2$} \\
\bottomrule
\end{tabular}
\vspace{2mm}
\caption{
Hyperparameters for fine-tuning \our{} on image-text retrieval.
}
\label{tbl:ft:retrieval:hyperparams}
\end{table}

\begin{table}[H]
\centering
\small
\begin{tabular}{l|c}
\toprule
\bf Hyperparameters & \bf ADE20K \\
\midrule
Peak learning rate & 1e-5 \\
Fine-tuning steps & 80k \\
Warmup steps & 1500 \\
Layer-wise learning rate decay & 0.95 \\
Batch size & 16 \\
AdamW $\epsilon$ & 1e-8  \\
AdamW $\beta$ & (0.9, 0.999) \\
Weight decay & 0.05 \\
Drop path & 0.5 \\
Dropout & \xmark \\
Input resolution & $896^2$ \\
\bottomrule
\end{tabular}
\vspace{2mm}
\caption{
Hyperparameters for fine-tuning \our{} on semantic segmentation.
}
\label{tbl:ft:semseg:hyperparams}
\end{table}

\begin{table}[H]
\centering
\small
\begin{tabular}{l|cc}
\toprule
\bf Hyperparameters & \bf  Object365 & \bf COCO \\
\midrule
Learning rate & 1e-4 & 5e-5 \\
Fine-tuning epochs & 15 & 20 \\
Warmup steps & \multicolumn{2}{c}{250} \\
Layer-wise learning rate decay & \multicolumn{2}{c}{0.9} \\
Batch size & \multicolumn{2}{c}{64} \\
AdamW $\epsilon$ & \multicolumn{2}{c}{1e-8}  \\
AdamW $\beta$ & \multicolumn{2}{c}{(0.9, 0.999)} \\
Weight decay & \multicolumn{2}{c}{0.1} \\
Drop path & \multicolumn{2}{c}{0.6} \\
Input resolution & $1024^2$ & $1280^2$ \\
\bottomrule
\end{tabular}
\vspace{2mm}
\caption{
Hyperparameters for fine-tuning \our{} on object detection.
}
\label{tbl:ft:od:hyperparams}
\end{table}

\begin{table}[H]
\centering
\small
\begin{tabular}{l|cc}
\toprule
\bf Hyperparameters & \bf ImageNet-21K & \bf ImageNet-1K \\
\midrule
Peak learning rate & 5e-5 & 3e-5 \\
Fine-tuning epochs & 50 & 15 \\
Warmup epochs & 5 & 3 \\
Layer-wise learning rate decay & 0.85 & 0.95 \\
Batch size & $16$k & $2$k \\
AdamW $\epsilon$ & 1e-6 & 1e-8  \\
AdamW $\beta$ & (0.9, 0.98) & (0.9, 0.999) \\
Weight decay & \multicolumn{2}{c}{0.05} \\
Drop path & \multicolumn{2}{c}{0.4} \\
Dropout & \multicolumn{2}{c}{\xmark} \\
Input resolution & $224^2$ & $336^2$ \\
Label smoothing $\varepsilon$ & \multicolumn{2}{c}{0.1} \\
\bottomrule
\end{tabular}
\vspace{2mm}
\caption{
Hyperparameters for fine-tuning \our{} on image classification.
}
\label{tbl:ft:imagenet:hyperparams}
\end{table}

\end{document}